# A Reconfigurable Mixed-signal Implementation of a Neuromorphic ADC

Ying Xu, Chetan Singh Thakur, Tara Julia Hamilton, Jonathan Tapson, Runchun Wang, André van Schaik
The MARCS Institute, University of Western Sydney, Sydney, NSW, Australia
ying.xu@uws.edu.au

*Abstract*— We present a neuromorphic Analogue-to-Digital Converter (ADC), which uses integrate-and-fire (I&F) neurons as the encoders of the analogue signal, with modulated inhibitions to decohere the neuronal spikes trains. The architecture consists of an analogue chip and a control module. The analogue chip comprises two scan chains and a two-dimensional integrate-and-fire neuronal array. Individual neurons are accessed via the chains one by one without any encoder decoder or arbiter. The control module is implemented on an FPGA (Field Programmable Gate Array), which sends scan enable signals to the scan chains and controls the inhibition for individual neurons. Since the control module is implemented on an FPGA, it can be easily reconfigured. Additionally, we propose a pulse width modulation methodology for the lateral inhibition, which makes use of different pulse widths indicating different strengths of inhibition for each individual neuron to decohere neuronal spikes. Software simulations in this paper tested the robustness of the proposed ADC architecture to fixed random noise. A circuit simulation using ten neurons shows the performance and the feasibility of the architecture.

## I. INTRODUCTION

Conventional ADCs encode continuous-time signals into a discrete-level representation. Their output data rate is independent of the signal characteristics. For example, if the input signal is zero, the ADC would continue to generate samples at a fixed output rate. For certain applications, some degradation can be tolerated in the recovery of the signal as long as the power consumption, size and bandwidth constraints are met. However, the conventional ADCs fail to utilise the signal information resulting in unnecessary power and bandwidth consumption. Owing to the shortcoming there has been a significant interest in exploring novel schemes for analogue-to-digital conversion. An alternative approach is to use irregular samplers such as neuromorphic ADCs [1]–[6]. For example, Watson and his colleagues have used single-layer integrate-and-fire networks with inhibitory connections for analogue-to-digital conversion in [5]. In their proposed asynchronous system, all the neurons are fully connected with the lateral inhibitions. These neuromorphic ADCs use integrate-and-fire neurons as time or rate encoders of the analogue input. Their advantages include speed, accuracy, and robustness to noise and circuit mismatch. Lateral inhibition implemented in these systems is used to decohere the parallel pathways. This prevents the neurons from firing at the same time, so that the neuronal spikes from different neurons can distribute evenly across time for a constant input value. However, in practice the same inhibitory connections to all neurons does not decohere the spikes well due to delays in the feedback (inhibitory) connections. In this paper, we propose a neuromorphic ADC architecture that makes use of an analogue two-dimensional integrate-and fire neuronal array as the basis, with a control module implemented on an FPGA to access individual neurons in the network, and a pulse width modulation methodology to control the inhibition connection among all the neurons. Each neuron receives different amounts of inhibition, which ensures decoherence of the neuronal spikes. The network architecture is shown in Fig. 1.

## II. ARCHITECUTRE

### A. Architecture

The programmable neuromorphic ADC architecture proposed in this paper consists of a control module and an analogue neuronal chip. The control module is called the inhibition generator, which is implemented on an FPGA. It comprises a scan enable generator and a pulse width modulator. The analogue neuronal chip consists of a two-dimensional array of integrate-and-fire neurons and shift registers. The architecture is shown in Fig. 2.

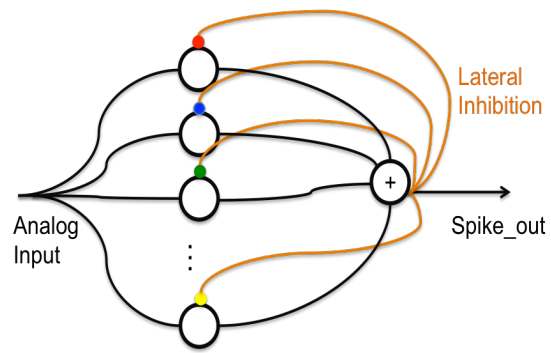

Fig. 1. ADC network; the analogue signal is fed into all the neurons in the network, and if one of the neurons fires, all neurons receive different strengths of inhibitions by using the pulse width modulation methodology.

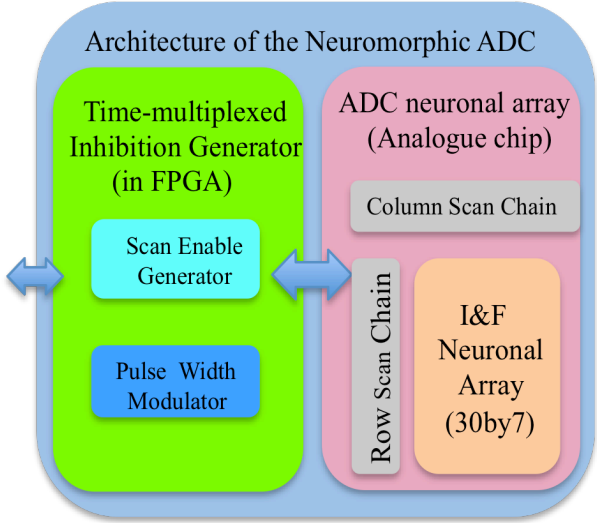

Fig. 2. Architecture of the Neuromorphic ADC.

The integrate-and-fire neurons are arranged in a two-dimensional array. Two scan enable pulses from the inhibition generator are shifted along the column and the row scan chain respectively. A neuron is selected if both its column and row enable pulse are reached. Hence, the interface between the inhibition generator and the chip becomes simple and hardware-friendly.

All the neurons use the same input port and they will leaky-integrate the analogue input signal all the time. The output channel and the inhibition channel are shared by all the neurons using a time-multiplexing approach [7]–[13]: only the selected neuron drives the output and receives the inhibition channel in the time step at which it is selected.

For example, as shown in Fig. 3 (A), one row and three columns of neurons are used to form a network. At the first time step T1, the scan enable signal reaches the first column register, and Neuron1 is selected. If the membrane voltage of Neuron1 reaches the threshold, it will generate a spike to the output channel, as well as an inhibition flag to the inhibition generator.

As the inhibition generator receives the flag, it will first check whether Neuron1 has received any inhibition signals from other neurons. If Neuron1 has not received any inhibitions from other neurons, the inhibition generator will send a series of inhibition signals with different pulse widths to Neuron1 (blue pulse in Fig. 3 (B)), Neuron2 (pink pulse) and Neuron3 (green pulse) sequentially at time step T1, T2 and T3. The first (blue) pulse width is the longest one to reset the membrane voltage of Neuron1 to zero. Otherwise, if Neuron1 has already received an inhibition pulse from another neuron, the inhibition generator will not generate any inhibitions anymore, and only a spike will be generated in T1.

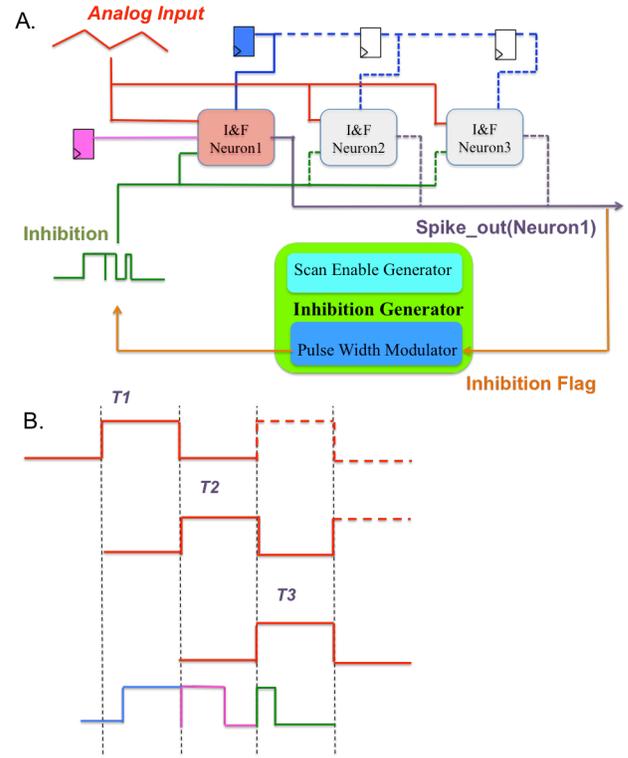

Fig. 3. (A) Time-multiplexing approach; (B) Pulse width modulation methodology.

If the membrane voltage of Neuron1 is below the threshold, neither the spike nor the inhibition flag is generated within the current time step. Then the next neuron Neuron2 in the next time step is scanned, and Neuron3 at the third time step T3 is followed. After all the three neurons have been scanned, their membrane voltages have been set to different levels. As a result their spikes will be decohered.

Additionally, the inhibition generator on the FPGA provides a programmable way for us to decide the inhibition algorithm to the neurons in the network. Any change and improvement of the inhibition methodology can be accomplished by simply changing the Verilog code on the FPGA.

*B. Two-dimensional Neuronal Array*

The analogue chip has been fabricated using IBM 130nm technology for prototyping. We have used low power transistors in the chip to achieve low power consumption. The chip consists of two scan chains and a 30by7 integrate-and-fire neuronal array. Thus, a total of 210 neurons distributed across 30 columns and 7 rows, constitute this two dimension neuronal array.

*C. Integrate-and-fire Neuron*

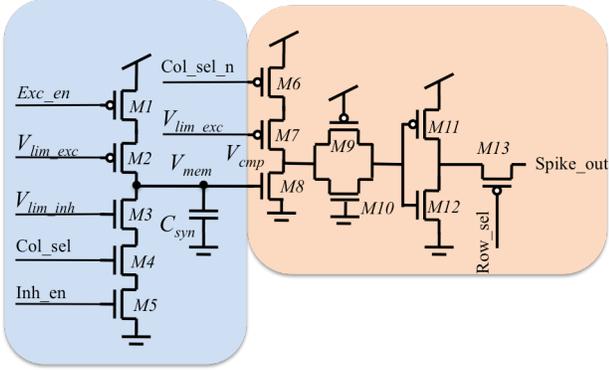

Fig. 4. Schematic of the Integrate-and-fire neuron.

The schematic of the integrate-and-fire neuron is shown in Fig. 4. For simplicity, we assume that the neuron is always being selected and all the transistors controlled by the selected signals are conducting. The analogue input current is mirrored to the node *Exc_en* via a current mirror; the node $V_{lim\_exc}$ is connected to the gate of M2 as well as M7. M2 is a cascade transistor to limit the source-drain voltage of M1. M7 is used in a current comparator, which compares the current through M6-M7 and the current through M8. The node $V_{lim\_inh}$ is utilised to limit the discharging current through M3. The node Inh_en indicates the inhibition signals, which are generated from the inhibition generator. The MOS capacitor, $C_{mem}$, is charged and discharged through M1-M2 and M3-M5 respectively. If the pull-up current through M6-M7 is larger than the pull-down current through M8, which is controlled by $V_{mem}$, $V_{cmp}$ is pulled to high; otherwise $V_{cmp}$ is pulled down to the ground. $V_{cmp}$ is transmitted to the output node through a transmission gate (M9-M10) and an inverter (M11-M12).

Additionally, if the neuron is not selected, the node Col_sel is low and the node Col_sel_n is high, there is no current path from $V_{dd}$ to ground, hence this design helps to reduce the power consumption of the circuit.

### III. SIMULATION RESULTS

#### A. Software Simulation

We have modelled our neuromorphic ADC with 50 neurons in MATLAB. We simulated it with a sinusoidal input (1μA peak to peak, with 2μA offset) and observed the spikes from the neuronal array in each time step ( Fig. 5 (A)).

In the software simulation, we have considered fixed random noise, which occurs due to random device mismatches of the transistors during the fabrication process. We have considered 20% mismatch in the input current mirror circuit (Fig. 4) and 30% mismatch in the inhibition control circuit.

In Fig. 5 (A), we can see the distribution of spikes with respect to the input signal. Fig. 5 (B) shows the reconstruction of the input signal with 6% RMS error with respect to the input signal. We have reconstructed the desired input signal from the spike counts at each time step by using a simple low pass filter, to demonstrate the neuromorphic ADC performance. We have used a systematic offset and a random component in the inhibition pulse to make a different inhibition for each neuron. We executed the simulation 30 times to check the robustness of the system for different randomness, and achieved a mean RMS error of 7.264% and a standard deviation of 1.362%.

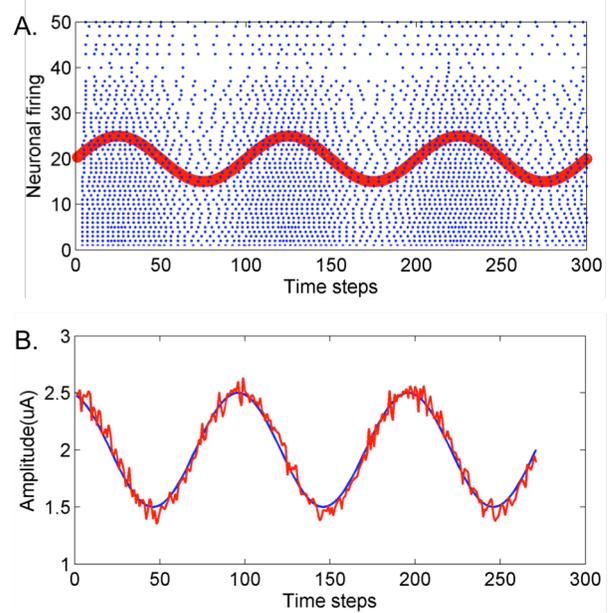

Fig. 5. (A) Firing distribution of 50 neurons corresponding (blue dots) to sinusoidal input (red curve); (B) Input signal is shown in blue and red curve shows the reconstructed signal from the neuronal spike trains.

#### B. Circuit Simulation

To test the performance of this neuromorphic ADC architecture and the pulse width modulation methodology, we have simulated the circuit using the IBM 130nm technology in Spectre. To reduce the simulation time, we have simulated one row by ten columns of neurons, totalling ten neurons. All the neurons use the same size in the spice model. In this simulation, a saw tooth waveform was used as the input signal, which is fed into *Exc_en*. One cycle of the input current is increased from 0 A at 0s to 100 nA at 25 μs, and back to 0 A at 50 μs (Green line in Fig. 6) The input $V_{lim\_exc}$ is controlled by a current mirror, which is set to 800 nA. The input $V_{lim\_inh}$ is controlled by a current mirror, which is set to 4 μA. The column shift rate is set to 33.3 MHz and the clock frequency of the inhibition generator is 333 MHz in the Verilog simulation. The output is the spike train from the ten neurons in different time steps. An average output spiking rate of 6.6 spikes/μs is achieved here (Blue lines in Fig. 6). The membrane voltage of each neuron is set to a different value in as the result of the pulse width modulation. Fig. 7 shows the reconstruction of the input analogue signal by using the neuronal spikes generated from the circuit simulation. We have presented input to the neural ADC as a ramp signal as shown in the red curve and observe the spike counts. We have applied curve fitting to the spike counts and obtained spikeCountNorm signal, which is normalised to its maximum value, shown in the black curve. We have calculated the compensation function by finding the inverse of the spikeCountNorm signal.

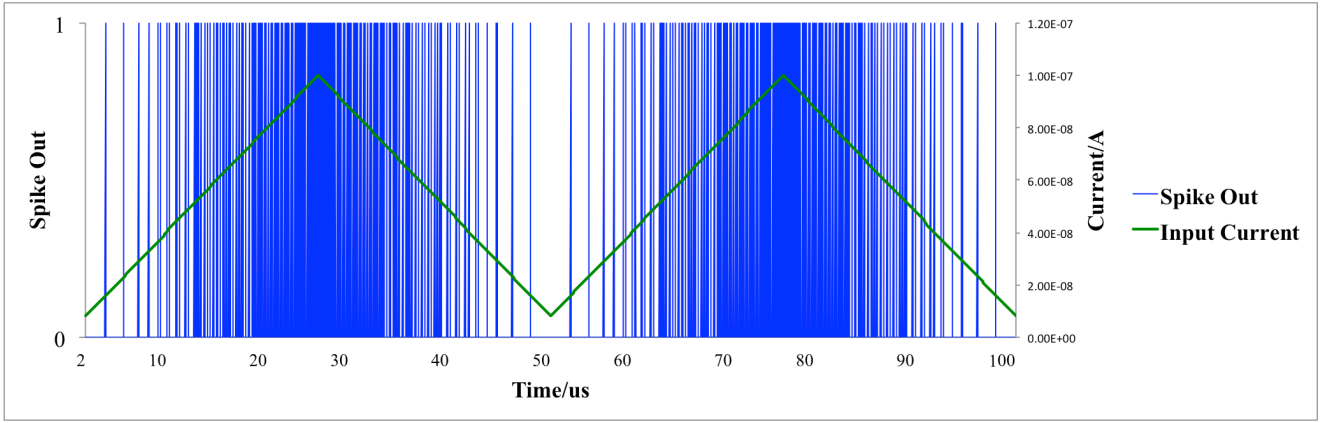

Fig. 6. Simulation Results; Input Current (green line) and Spike Out (blue pluses).

## IV. CONCLUSIONS

We have presented a reconfigurable mixed-signal implementation of a neuromorphic ADC, which utilises FPGA implemented control module to scan the two-dimensional integrate-and-fire neuronal array, with a pulse width modulation algorithm to decohere the neuronal spikes. The software simulation results show satisfactory performance and circuit simulations using ten neurons show the even decoherence of the spikes from the ten neurons. It proves the feasibility of the architecture for the integrated circuit design.

## V. ACKNOWLEDGMENT

This work has been supported by the Australian Research Council Grant DP140103001. The support by the Altera and Xilinx university program is gratefully acknowledged. This work was inspired by the Capo Caccia Cognitive Neuromorphic Engineering Workshop 2014 and Telluride Neuromorphic workshop 2014.

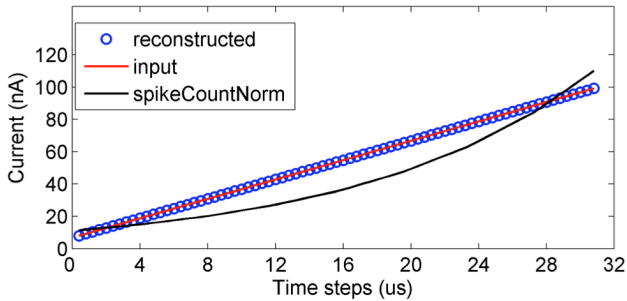

Fig. 7. Reconstruction; Input to the neural ADC (red curve); Curve fitting for normalised spike counts (black curve); reconstructed signal (blue circle).